\documentclass[letterpaper, 10 pt, conference]{ieeeconf}  

\IEEEoverridecommandlockouts                              

\let\labelindent\relax

\usepackage{graphics}           
\usepackage{times}              
\usepackage{amsmath}            
\usepackage{amssymb}            
\usepackage{graphicx}
\usepackage{algorithm}
\usepackage[noend]{algpseudocode}
\usepackage{booktabs}
\usepackage{color}
\definecolor{instructioncolor}{rgb}{.5,.5,.5}

\usepackage[font=small]{caption}

\def\figref#1{Fig.~\ref{#1}}

\def\eqref#1{Eq.~(\ref{#1})}

\makeatletter
\usepackage{xspace}
\DeclareRobustCommand\onedot{\futurelet\@let@token\@onedot}
\def\@onedot{\ifx\@let@token.\else.\null\fi\xspace}
\def\eg{e.g\onedot}

\makeatother

\usepackage{array}
\newcolumntype{L}[1]{>{\raggedright\let\newline\\\arraybackslash\hspace{0pt}}m{#1}}
\newcolumntype{C}[1]{>{\centering\let\newline\\\arraybackslash\hspace{0pt}}m{#1}}
\newcolumntype{R}[1]{>{\raggedleft\let\newline\\\arraybackslash\hspace{0pt}}m{#1}}

\def\argmin{\mathop{\rm argmin}}

\newcommand{\norm}[1]{\lVert#1\lVert}

\renewcommand{\v}[1]{{\b #1}} 

\usepackage[T1]{fontenc}      
\usepackage{lmodern}          
\usepackage[utf8]{inputenc}   

\usepackage{graphicx}
\usepackage{booktabs}
\usepackage{xcolor}

\usepackage{cite}
\usepackage{epsfig}
\usepackage{multirow}
\usepackage{makecell}
\usepackage{amsfonts}
\usepackage{enumitem}
\usepackage{array}
\usepackage{algorithm}
\usepackage{algpseudocode}
\usepackage[symbol]{footmisc}
\usepackage[accsupp]{axessibility}
\usepackage{blindtext}
\usepackage{graphicx}
\usepackage{mathtools}
\usepackage{arydshln}
\usepackage{graphicx}
\usepackage{subcaption}

\usepackage{tabularx}
\usepackage{adjustbox}
\usepackage{multirow}
\usepackage{booktabs}
\usepackage{float} 

\usepackage{colortbl}
\usepackage{amssymb}
\usepackage{pifont}

\algrenewcommand\algorithmicindent{0.75em}

\def\eqref#1{(\ref{eq:#1})}
\def\eqlabel#1{\label{eq:#1}}
\def\figref#1{\ref{fig:#1}}
\def\figlabel#1{\label{fig:#1}}
\def\pparagraph#1{\par{\bf #1}~~}

\def\eqref#1{(\ref{eq:#1})}
\def\eqlabel#1{\label{eq:#1}}
\def\figref#1{\ref{fig:#1}}
\def\figlabel#1{\label{fig:#1}}
\def\pparagraph#1{\par{\bf #1}~~}

\def\xcomment#1{\textcolor[rgb]{.3,.3,.1}{\text{$/\!\!/$ {\em #1}}}}
\def\comment#1{\kern-1cm\xcomment{#1}}
\def\eqcomment#1{\kern-1cm\xcomment{#1}}

\def\v#1{\ensuremath{\mathbf{#1}}}

\def\real{\mathbb{R}}

\def\norm#1{\left\lVert#1\right\rVert}

\def\l2#1{\norm{#1}_2}

\usepackage{graphics} 
\usepackage{epsfig} 
\usepackage{mathptmx} 
\usepackage{times} 
\usepackage{amsmath} 
\usepackage{amssymb}  

\makeatletter
\long\def\@makefntext#1{%
  \parindent 0pt%
  \noindent
  #1%
}
\makeatother

\title{\LARGE \bf ReLoc: Rethinking Scene Coordinate Regression Architecture for Robust Outdoor LiDAR-based Localization}

\author{Heejoon Moon$^{1}$ \and Yurim Cho$^{2}$ \and Je Hyeong Hong$^{*1,2}$
\thanks{* Corresponding author: Je Hyeong Hong}%
\thanks{$^{1}$Department of Artificial Intelligence, Hanyang University}
\thanks{$^{2}$Department of Electronic Engineering, Hanyang University}
\thanks{E-Mail: \{wilko97,csacyr,jhh37\}@hanyang.ac.kr}
\thanks{Code is available at https://github.com/SpatialAILab/ReLoc}
\thanks{
This work was supported by National Research Foundation of Korea (NRF) grants funded by the Korea government (MSIT) (No.RS-2026-25486241, 50\%; No.RS-2025-16068784, 25\%) and Commercialization Promotion Agency for R\&D Outcomes grants funded by the Korea government (MSIT) (No.RS-2026-25551924, 25\%).
}
}

\begin{document}
\maketitle
\thispagestyle{empty}
\pagestyle{empty}

\begin{abstract}

Scene Coordinate Regression (SCR) has recently emerged as a promising approach for LiDAR-based localization, achieving accurate localization without requiring an explicit 3D map. Despite their effectiveness, existing SCR methods rely on scene classification-based global embedding that struggles to provide fine-grained discrimination among nearby locations. Moreover, their reliance on uniform sampling of local features during training assigns equal importance to all points, thereby inadvertently propagating features from dynamic objects or unstable regions and potentially degrading training stability. In this paper, we present ReLoc, a revamped SCR architecture that can effectively address these limitations. First, we redesign the global embedding module by combining learnable context tokens with a feature aggregator to capture richer and more discriminative scene context. Second, we introduce an attention-based local feature enhancement module to mitigate the impact of noisy local features while encouraging context-consistent structures, yielding more robust local feature representations. Experimental results on two large-scale outdoor datasets demonstrate that our approach achieves state-of-the-art accuracy over previous SCR-based methods while maintaining real-time inference performance.
\end{abstract}

\section{INTRODUCTION}
\label{sec:introduction}

LiDAR-based localization is crucial in autonomous systems, including self-driving vehicles~\cite{Yin_2024_IJCV} and mobile robotics~\cite{Zhang_2024_AAAI}. 
A common paradigm is structure-based localization, where a 6-DoF pose is estimated by aligning incoming LiDAR scans with a pre-built 3D map. 
This is typically achieved by establishing point correspondences using geometric descriptors~\cite{FPFH, GEDI, FCGF, DIP, GCL, SPINNET}, followed by pose estimation through Procrustes alignment~\cite{Procrustes} equipped with RANSAC~\cite{RANSAC}.
Despite achieving precise localization using explicit 3D maps with stored descriptors,
this paradigm introduces several limitations. Specifically, it incurs i) substantial storage demands for 3D points \& descriptors, ii) increases communication overhead for large-scale maps, and iii) raises privacy concerns during data transmission~\cite{pittaluga2019revealing,song2020deep}.

Recently, learning-based LiDAR localization methods have emerged as a promising alternative by eliminating the need for prior 3D maps.
By encoding the information of prebuilt 3D maps implicitly within network parameters, these methods effectively alleviate storage and privacy concerns while allowing 6-DoF pose estimation directly from LiDAR scans.
Remarkably, scene coordinate regression (SCR) methods~\cite{SGLoc, LiSA, LightLoc, GTRLoc} have achieved in-meter localization accuracy by predicting per-point correspondences between a LiDAR scan and map coordinates.
Then, these predicted 3D-3D correspondences are used to estimate the 6-DoF pose via robust solvers like RANSAC~\cite{RANSAC}, enhancing robustness to outliers.
Compared to the methods that directly regress the pose from the query LiDAR scan~\cite{STCLoc, NIDALoc, HypLiLoc, DiffLoc}, SCR-based approaches benefit from explicit (point-wise) geometric supervision, thereby achieving accurate localization.

\begin{figure}[t]
\centering
    \subfloat{
    \includegraphics[width=0.48\linewidth]{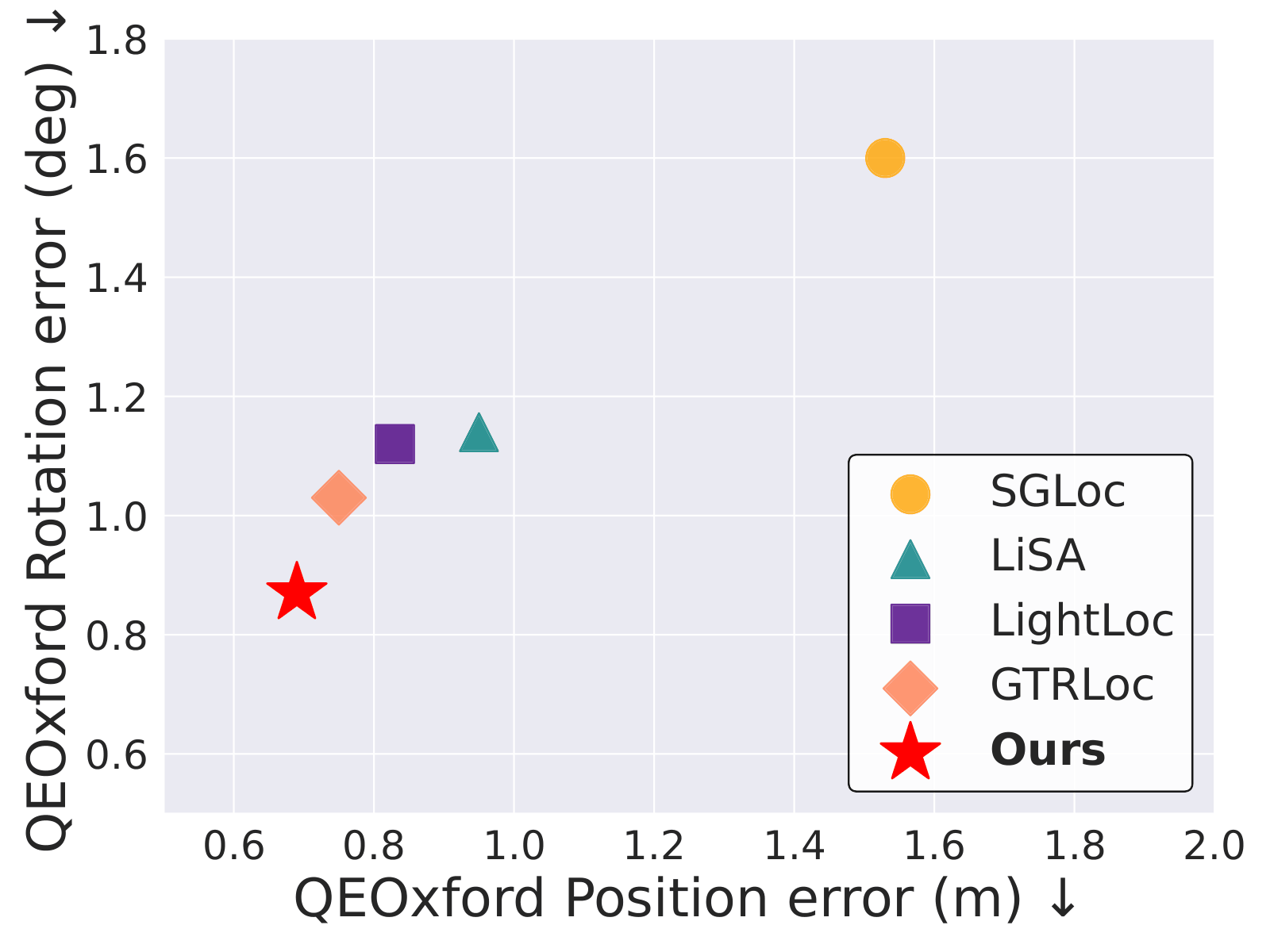}
    \includegraphics[width=0.48\linewidth]{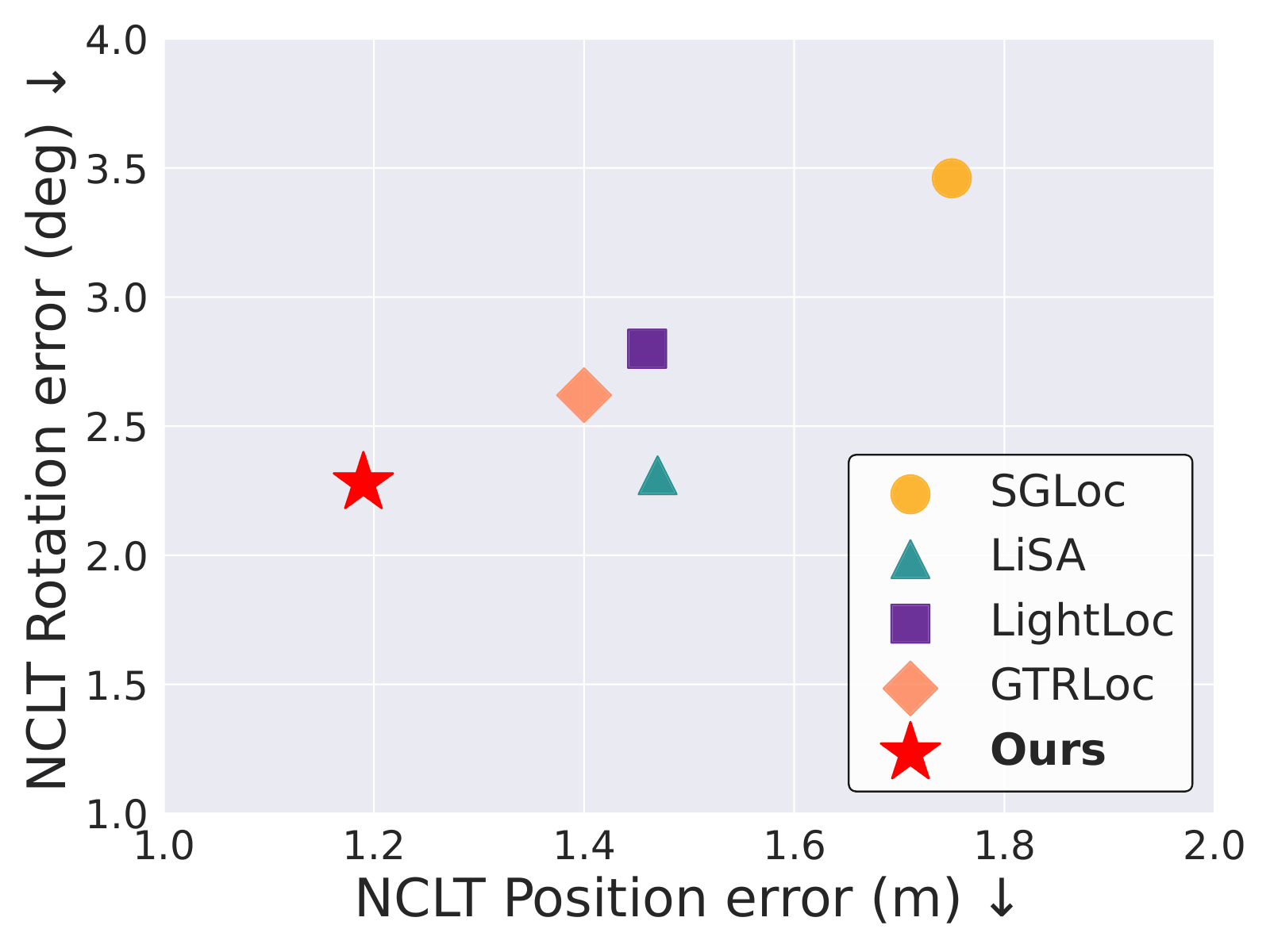}
    }
\caption{\textbf{Comparison of localization accuracy across recent LiDAR-based SCR methods}. Our method achieves the lowest translation and rotation errors on the QEOxford and NCLT datasets while maintaining real-time inference capability of over 90Hz.}
\vspace{-4mm}
\label{fig:teaser}
\end{figure}

Recent SCR methods~\cite{LightLoc, GTRLoc} commonly adopt hierarchical architectures that combine global and local embeddings for accurate large-scale localization.
Nevertheless, we show such designs contain potential limitations. 
First, for extracting global embeddings, existing methods rely on scene classification logits from the global average-pooled vector of local features of the input query scan, which may contain features from dynamic objects and unstable regions.
Also, since classification logits are primarily optimized for coarse scene discrimination, they lack fine-grained details required for distinguishing samples within the same scene cluster (see Fig.~\figref{global_analysis}). 
Second, regarding local embeddings, prior studies~\cite{SGLoc,LightLoc,GTRLoc} employ uniform sampling of features during training, which inevitably includes noisy features from dynamic objects or unstable regions inconsistently observed across multiple sequences.
As shown in Fig.~\figref{lightloc_analysis}, these dynamic/unstable features tend to yield higher regression errors and lower inlier ratios, suggesting that they can adversely influence training stability and localization.

To address above limitations, we present \textbf{ReLoc}, a revamped SCR architecture that replaces conventional global and local embedding modules with our proposed method.
First, we introduce a \textit{scene context-aware global embedding module} that combines learnable context tokens with an MLP-Mixer~\cite{tolstikhin2021mlp}-based feature aggregator.
The learnable tokens enable richer scene context representation, while the feature aggregator efficiently aggregates these tokens into a compact global embedding.
This allows the network to capture highly discriminative scene-level context, addressing the lack of fine-grained discrimination within the same cluster in existing classification-based embeddings.
Second, we introduce an attention-based \textit{local feature enhancement module}. 
By refining local features through residual connections, the attention mechanism mitigates noisy features and promotes context-consistent interactions, providing more reliable representations for accurate coordinate regression.
Extensive experiments on two large-scale outdoor datasets demonstrate that both modules work synergistically to improve LiDAR localization accuracy beyond the current state-of-the-art SCR methods while maintaining real-time performance.

\begin{figure}[t]
    \centering
    \vspace{2mm}
    \includegraphics[width=0.9\linewidth]{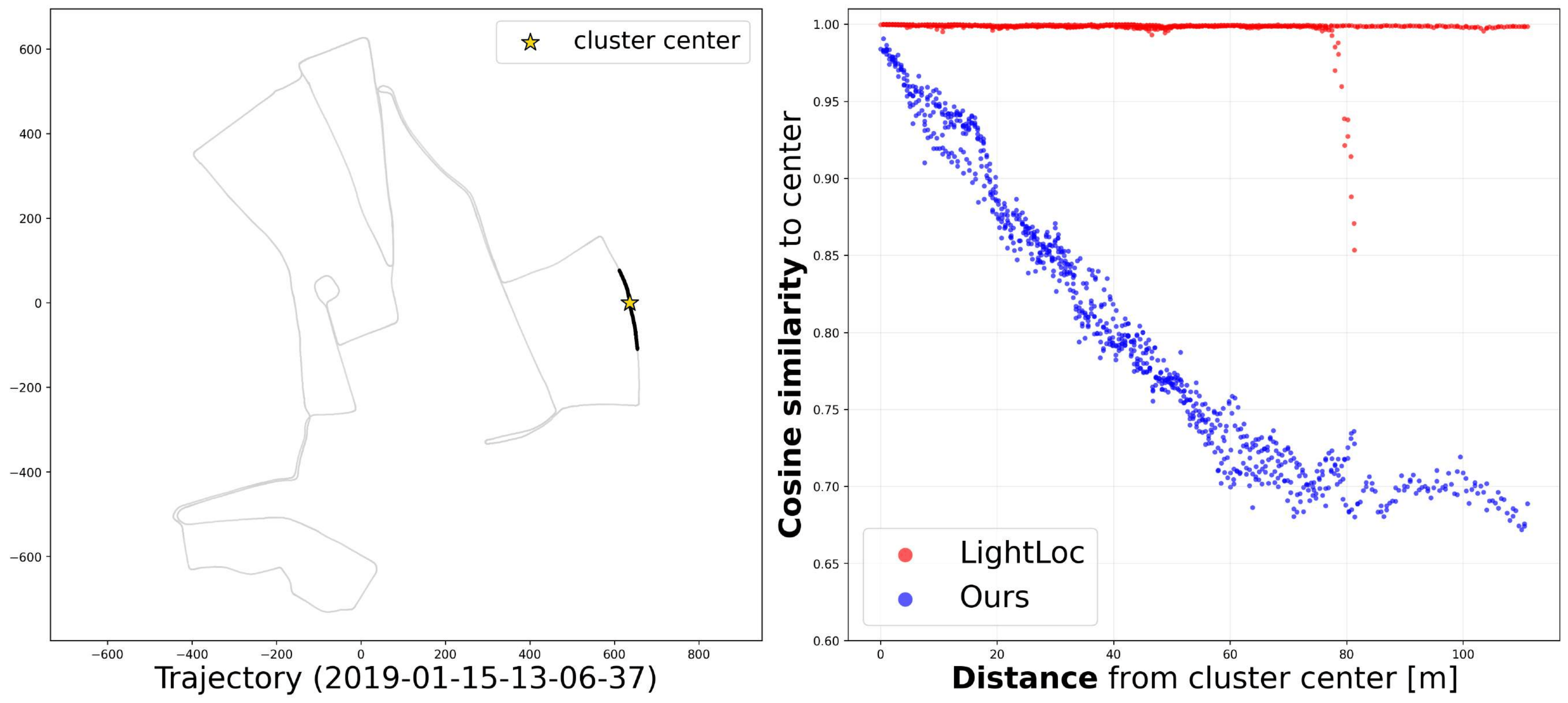}
    \caption{
    \textbf{Comparison of global embeddings.} We evaluate the discriminative ability of global embeddings within the same cluster (colored \textbf{black} on left) by measuring their cosine similarity to a scene cluster center ($\star$) on the \texttt{15-13-06-37} in the QEOxford dataset. Compared to classification logits-based embeddings (LightLoc~\cite{LightLoc}) that lack fine-grained discrimination by maintaining erroneously high similarity for spatially distant samples, our proposed global embedding provides more discriminative information to the SCR model, yielding high cosine similarity for nearby scans while naturally decreasing as the spatial distance increases. 
    }
    \figlabel{global_analysis}
    \vspace{-5mm}
\end{figure}

\section{Related Work}
\label{sec:related_works}

\subsection{Structure-based LiDAR localization}
Structure-based LiDAR localization aims to estimate the 6-DoF pose of a query scan by exploiting geometric and structural information from a pre-built 3D map. 
A common paradigm is to register the input scan against a prior map by establishing per-point correspondences using learned or handcrafted descriptors~\cite{FPFH, GEDI, GCL}, enabling accurate pose estimation. 
Structure-based localization methods can be broadly categorized into retrieval-based~\cite{Yu_2021_ISPRS, Xia_2021_CVPR, Xia_2023_ICCV, Wang_2024_AAAI, Luo_2023_ICCV} and registration-based approaches~\cite{Zhang_2023_CVPR, Jin_2024_CVPR, yang2020teaser, cui2024sage}. 
Retrieval-based methods formulate localization as place recognition by searching a database for similar scans, while registration-based methods explicitly estimate poses via descriptor matching followed by coarse-to-fine optimization, combining global estimators such as RANSAC~\cite{RANSAC} with local refinement methods like ICP~\cite{besl1992method}. 
However, these approaches rely on explicit map storage and incur memory overhead.

\begin{figure}[t]
    \centering
    \vspace{2mm}
    \includegraphics[width=0.9\linewidth]{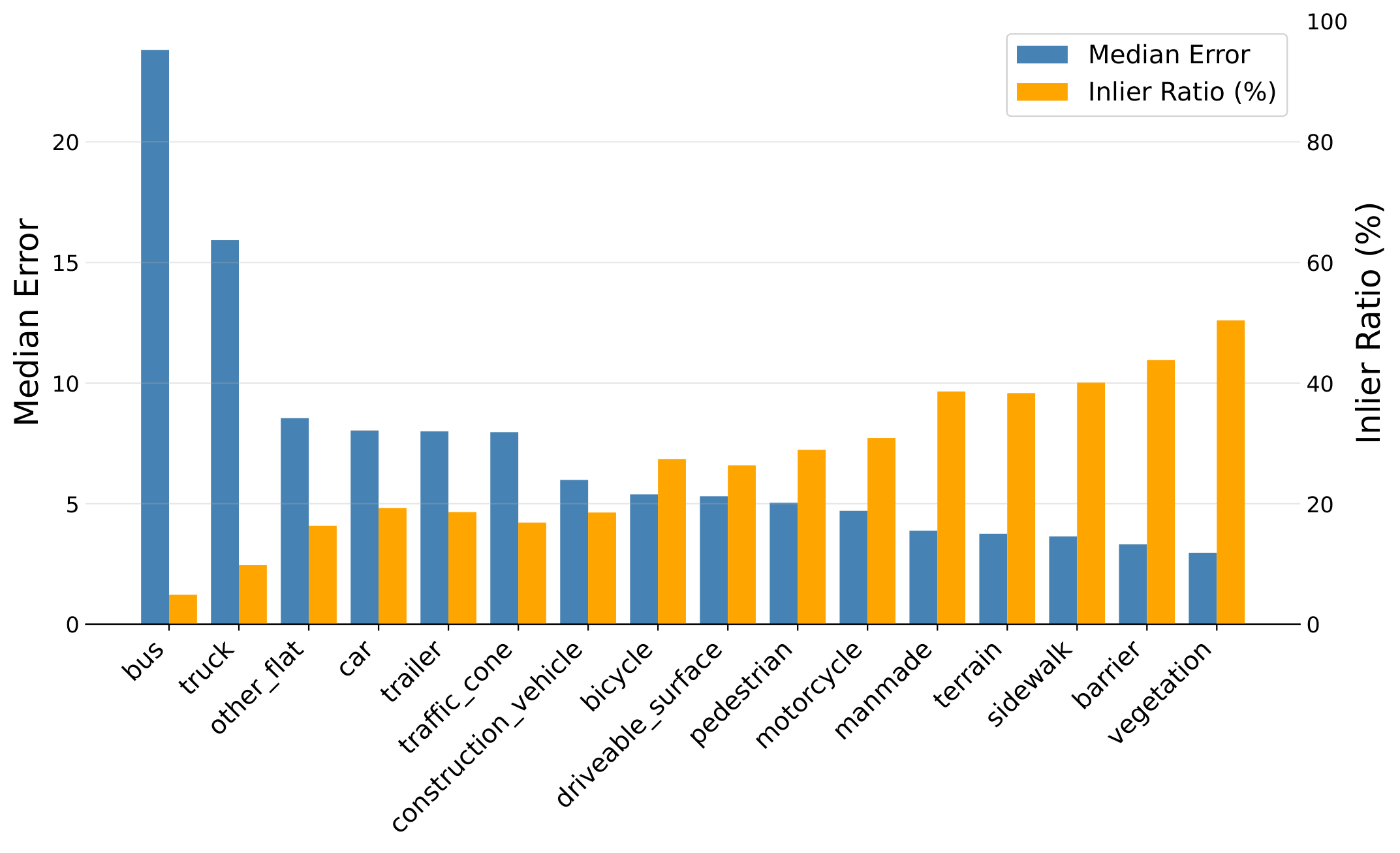}
    \caption{
    \textbf{Analysis of LightLoc}~\cite{LightLoc}. We analyze the relationship between the regression error \& inlier ratio and semantic meaning in the \texttt{15-13-06-37} sequence of the QEOxford.
    To categorize the query points according to semantic meaning, we employ a pretrained SphereFormer~\cite{SphereFormer}.
    Features from the dynamic categories (\eg vehicles, pedestrians) suffer from higher errors and lower inlier ratios, whereas features from static regions show the opposite trend.
    }
    \figlabel{lightloc_analysis}
    \vspace{-5mm}
\end{figure}

\subsection{Absolute pose regression}

LiDAR Absolute Pose Regression (APR) methods directly estimate a 6-DoF pose from input query LiDAR scan (3D point cloud) using a feed-forward network. Beyond early CNN-based architectures~\cite{PointLoc, Yu_2022_PR}, subsequent works imposed structural constraints via spatio-temporal aggregation~\cite{STCLoc} or bio-inspired memory~\cite{NIDALoc}, and enhanced representation capacity through Euclidean-hyperbolic feature fusion~\cite{HypLiLoc}.
More recently, DiffLoc~\cite{DiffLoc} and BEVDiffLoc~\cite{BEVDiffLocEL} adopt diffusion-based refinement for iterative pose estimation, and FlashMix~\cite{FlashMix} improves training efficiency via a scene-agnostic backbone and descriptor aggregator. 
Despite these advances, APR methods still struggle to achieve meter-level accuracy in complex outdoor environments.

While FlashMix is closely related to our study in terms of employing a MLP-Mixer aggregator, FlashMix relies on farthest point sampling, thereby including unstable dynamic elements. In contrast, we utilize learnable context tokens, which have been shown to effectively capture scene context by focusing on geometrically consistent structures~\cite{goswami2024salsa}.

\begin{figure*}[t]
    \centering
    \vspace{2mm}
    \includegraphics[width=0.95\linewidth]{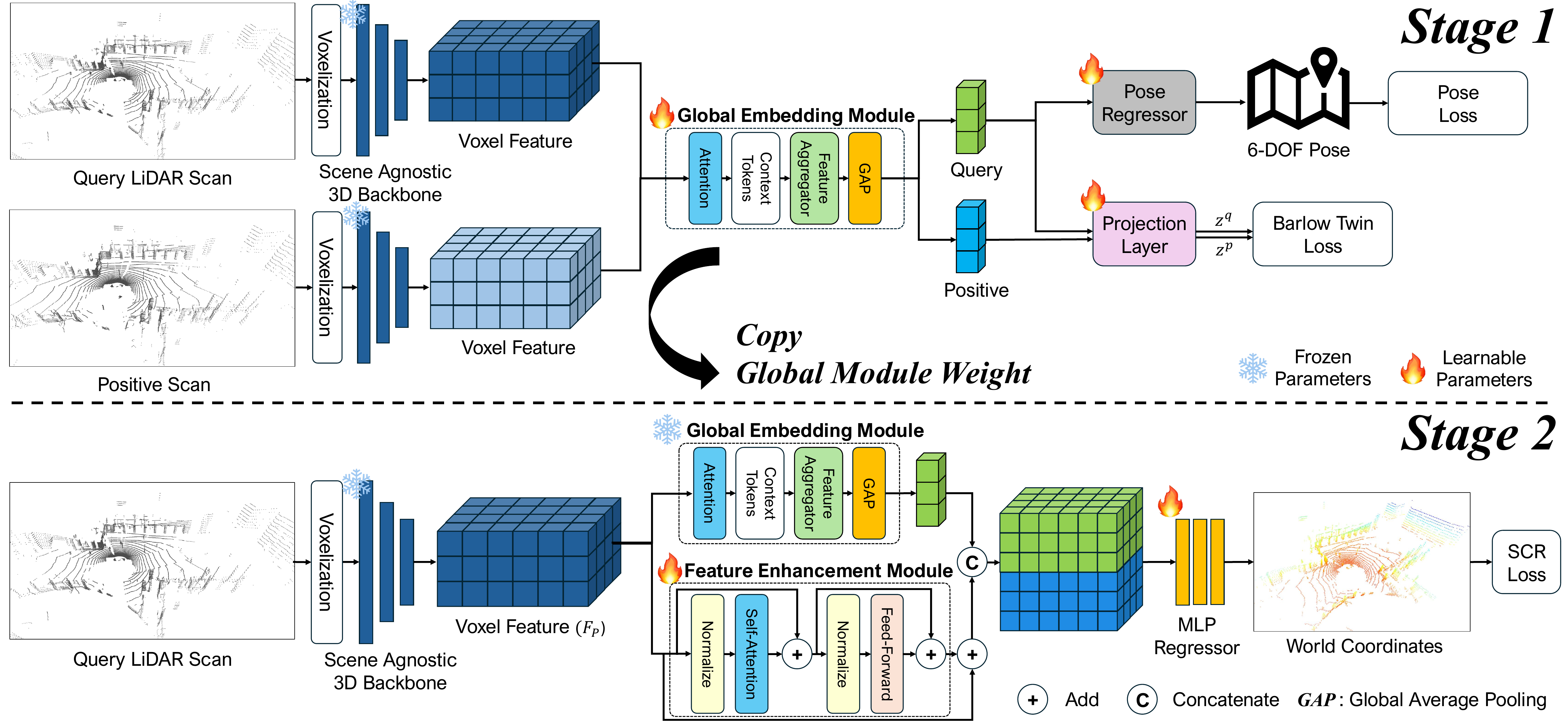}
    \caption{
    \textbf{Overview of our ReLoc framework.} 
    We adopt a two-stage training strategy. In Stage~1, the global embedding module is trained using 6-DOF pose supervision and contrastive regularization to learn context-aware scene representations. The learned weights are then transferred to Stage~2, where the global embeddings are concatenated with enhanced features through the Local Feature Enhancement Module, and an MLP-based regressor is trained to predict scene coordinates.
    }
    \figlabel{framework}
    \vspace{-7mm}
\end{figure*}

\subsection{Scene coordinate regression}
LiDAR-based SCR methods regress the world coordinates of input 3D points from the query input point cloud in LiDAR coordinates to establish 3D-3D correspondences.
The pose is then estimated by solving the Procrustes problem via RANSAC~\cite{RANSAC}.
SGLoc~\cite{SGLoc} firstly introduced the SCR pipeline into LiDAR-based localization.
It employs a 3D CNN-based feature extractor combined with a multi-scale feature aggregation module to encode scene geometry and regress scene coordinates. 
LightLoc~\cite{LightLoc} significantly reduces training time by employing a scene-agnostic feature extractor, coupled with a lightweight scene-specific MLP regression head.
GTRLoc~\cite{GTRLoc} further incorporates geospatial information to alleviate ambiguities arising from structurally similar scenes.
Other works address additional challenges such as rotation sensitivity~\cite{RALoc}, or limited training data~\cite{Unleashing}.
Despite their effectiveness, these methods exhibit limitations (details in Sec.~\ref{sec:preliminaires}) in both global and local representations. First, they rely on classification-based global embeddings that struggle to provide fine-grained discrimination among nearby locations. Second, they sample local features uniformly from the input scan during training, including noisy dynamic objects or unstable regions. While LiSA~\cite{LiSA} partly addresses the latter issue by incorporating semantic awareness, its performance relies on prior segmentation quality.

\section{Review of scene coordinate regression}
\label{sec:preliminaires}
LiDAR-based SCR framework aims to regress a set of 3D-3D correspondences $\mathbb{C} = \{(\v p_{i}^s, \v p_{i}^w)\}$, where $\v p_i^s\in \real^3$ denotes a 3D point in the voxelized query LiDAR scan $S$ and $\v p_i^w \in \real^3$ is the corresponding 3D point in the world coordinate frame.
Given the 3D-3D correspondences $\mathbb{C}$, the robust LiDAR-to-World transformation matrix $T_{w\leftarrow s} := [R|\v t] \in SE(3)$ can be estimated using an efficient solver for point cloud registration~\cite{sc2pcr} with RANSAC~\cite{RANSAC} as below
\begin{align}
T_{w\leftarrow s} =  \argmin_{[R\,|\,\v t]}  \sum_{i=1}^{M} \left\| R \v p_i^s + \v t - \v p_i^w \right\|_2^2,
\eqlabel{eq:ransac}
\end{align}
where $M$ denotes the number of correspondences.

To obtain the correspondences $\mathbb{C}$, SCR learns a mapping from LiDAR points to their corresponding world coordinates:
\begin{equation}
\v p_i^{w} = f\left(\v p_i^{s}; \mathbf{w}\right),
\eqlabel{eq:scr}
\end{equation}
where $f$ is a neural network parameterized by weights $\mathbf{w}$. 
The network $f$ is usually composed of a feature extractor ($f_B$) followed by a scene-specific MLP regression head ($f_H$).

\begin{figure*}[t]
    \centering
    \vspace{2mm}
    \includegraphics[width=0.95\linewidth]{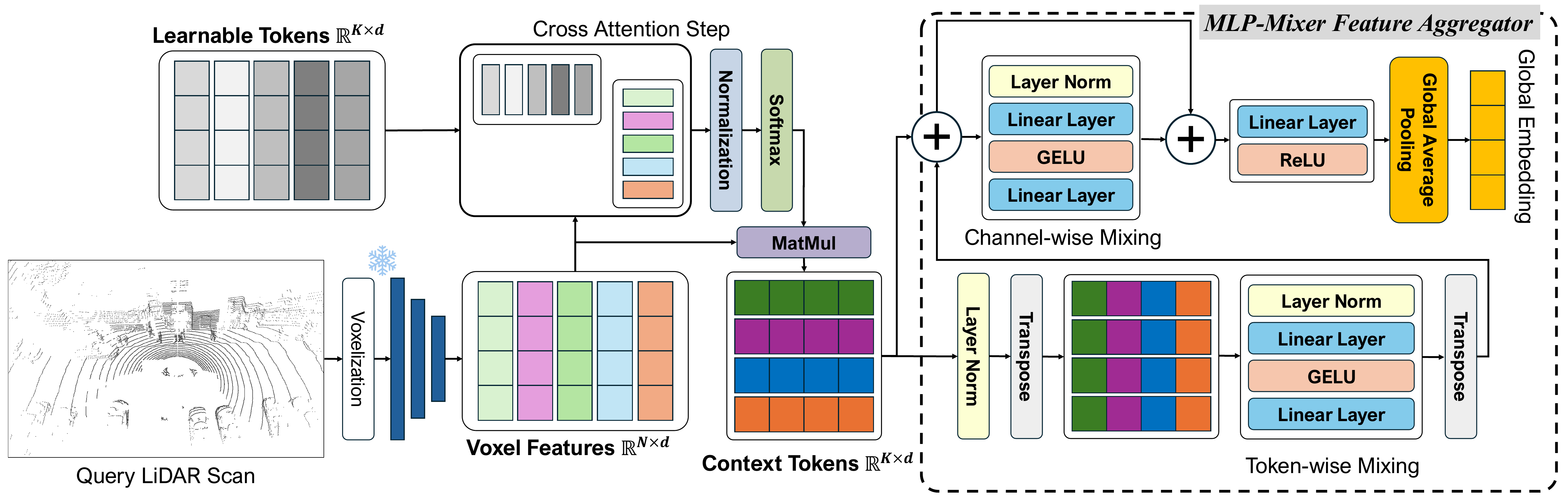}
    \caption{
    \textbf{Architecture of the global embedding module.} Learnable context tokens interact with voxel features via cross-attention to capture scene-aware context and are then aggregated by an MLP-Mixer to produce a compact global embedding for downstream regression.
    }
    \figlabel{global}
    \vspace{-6mm}
\end{figure*}

\pparagraph{Existing limitations.}
\label{sec:lightloc_analysis}
To identify the bottlenecks of current SCR methods, we evaluate the global and local representations of LightLoc~\cite{LightLoc}. 
First, by measuring the cosine similarity of global embeddings to a scene cluster center (shown in Fig.~\figref{global_analysis}), we observe that classification-based global embeddings lack fine-grained discrimination within the same cluster. By maintaining erroneously high similarity for spatially distant samples, they introduce ambiguity that hinders the SCR model from learning accurate coordinates.
Second, by analyzing the semantic meaning of query features (Fig.~\figref{lightloc_analysis}) during the inference, we observe that dynamic objects (\eg, vehicles) tend to yield higher regression errors and lower inlier ratios than static categories (\eg, manmade).
Our analysis suggests that SCR can be improved by capturing more discriminative global context and refining local representations to actively reinforce static structures while attenuating dynamic noise during training.

\section{Proposed Method}
\label{sec:method}

\pparagraph{Overview.} In this section, we present the scene context-aware global embedding module and the local feature enhancement module.
As shown in Fig.~\figref{framework}, our framework adopts a two-stage training strategy: the global embedding module ($f_G$) is first trained to learn context-aware global embedding and then fixed while the feature enhancement module ($f_E$) and scene-specific MLP regressor ($f_H$) are trained in the next step.
In Stage~1, context tokens are obtained via cross-attention with learnable tokens and local features ($F_P \in \real^{N\times d}$) to be aware of scan-specific context. 
These tokens are aggregated by a feature aggregator to produce a global embedding, passing the pose regressor to predict 6-DOF pose while also passing a projection head to induce similar embeddings for the positive scans (captured at the same location with a different sequence).
In stage~2, the local features ($F_P$) from the scene-agnostic backbone ($F_B$) are fed into the trained global embedding module and the local feature enhancement module to obtain a global representation and refined local features, respectively. These are then concatenated to form hierarchical embeddings and fed into an MLP-based regressor to predict the world coordinates.

\subsection{Scene context-aware global embedding module}
\label{sec:mixer_global}

Now, we introduce a detailed architecture of a global embedding module consisting of a pre-trained backbone, a scene context-aware attention mechanism, and a feature aggregator, as illustrated in Fig.~\figref{global}. 
We employ the scene-agnostic feature encoder~\cite{LightLoc} pre-trained on the nuScenes dataset~\cite{nuScenes} to extract the voxel-wise features ($F_P \in \real^{N\times d}$).
Then, to obtain a global representation,
we extract $K$ context tokens ($T_{\text{context}}\in\real^{K\times d}$) designed to preserve sub-region geometric structures.
Specifically, we obtain these context tokens through an attention mechanism, utilizing a set of learnable tokens as queries ($T_{\text{Learnable}}\in\real^{K\times d}$) while the local features $F_P$ serve as key and value,
\begin{equation}
T_{\text{context}} = \text{Attention}(Q={T_{\text{Learnable}}}, K=V=F_P).
\label{eq:adaptive_attention_pooling}
\end{equation}

To aggregate the $K$ context tokens ($T_{\text{context}}$, where $K$=128) into a single global representation, we process them using an MLP-Mixer-based feature aggregator~\cite{tolstikhin2021mlp,goswami2024salsa,FlashMix}, which alternates between token-wise and channel-wise mixing.
The token-mixing MLP enables long-range interactions across different spatial regions, allowing the model to reason over the global layout of the scene, while the channel-mixing MLP refines feature correlations within each token to enhance discriminative capacity.
Through this dual mixing process, the aggregator effectively captures both inter-token relationships and channel-level dependencies, leading to a more expressive and context-aware global representation.
Finally, a linear projection followed by global average pooling produces a single global descriptor per scan.

The obtained global descriptor is then fed into a pose regression head composed of stacked MLP layers, each consisting of a linear layer followed by batch normalization and a ReLU activation.
Through the regressor, 3D position ($\mathbf{t}_{pred} \in \mathbb{R}^3$) and the orientation in logarithmic quaternion form ($\mathbf{q}_{pred} \in \mathbb{R}^3$) are predicted.
Inspired by~\cite{FlashMix}, we further feed the global descriptor into a lightweight projection layer and apply contrastive regularization~\cite{barlow} to encourage consistent global representations across the positive scans.
    
\pparagraph{Loss function.} The training objective for the global embedding module consists of pose loss and the contrastive regularization term (Barlow Twins loss~\cite{barlow}) following~\cite{FlashMix}, 
\begin{equation}
L_{global} = L_{pose} + \lambda L_{Barlow}.
\eqlabel{eq:global_loss}
\end{equation}
The pose loss is computed as the sum of $L_1$-norm loss of predicted translation ($\textbf{t}^{pred}$) and orientation ($\textbf{q}^{pred}$),
\begin{equation}
L_{pose} = \left\| \textbf{t}^{pred} - \textbf{t}^{GT} \right\|_1 + \alpha \left\| \textbf{q}^{pred} - \textbf{q}^{GT} \right\|_1,
\eqlabel{eq:pose_loss}
\end{equation}
and the Barlow Twins loss is calcuated as 
\begin{equation}
L_{Barlow} = \sum_i (1 - C_{ii})^2 + \mu \sum_i \sum_{j \ne i} C_{ij}^2,
\eqlabel{eq:Barlow_loss}
\end{equation}
where $C$ denotes the cross-correlation matrix element between the projected embeddings of the query ($z^{q}$) and positive ($z^{p}$) scans across the batch, computed as $C_{ij} = \sum_{b} z_{b,i}^{q} z_{b,j}^{p} / \left( \sqrt{\sum_{b} (z_{b,i}^{q})^2} \sqrt{\sum_{b} (z_{b,j}^{p})^2} \right)$ with $b$ indexing the batch samples and $i, j$ denoting the feature dimensions of the projected embeddings.
The hyperparameters are set to $\lambda = 1.0e^{-4}$, $\alpha = 100$, and $\mu = 5.0e^{-3}$.

\begin{figure}[t]
\centering
\includegraphics[width=0.96\linewidth]{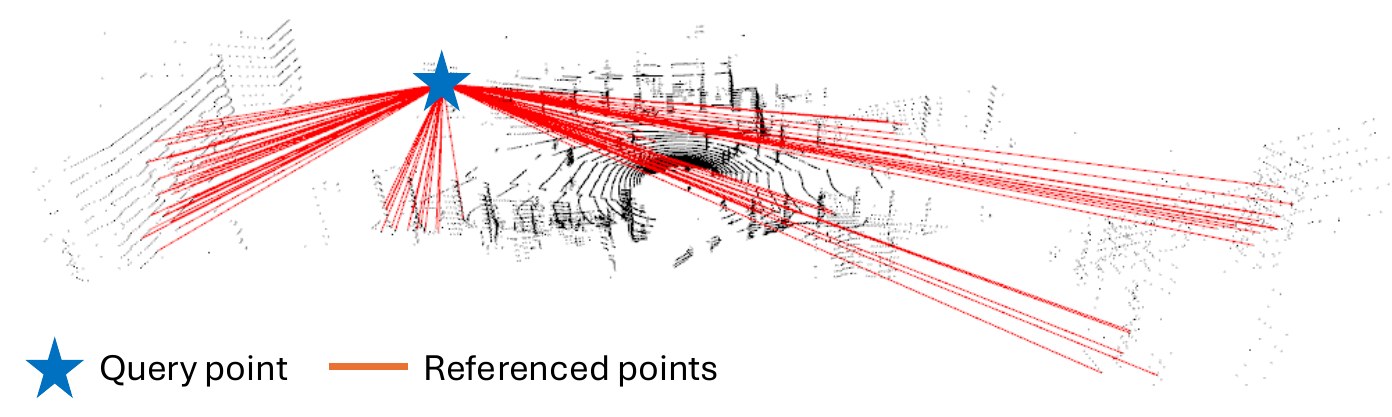}
\vspace{-2mm}
\caption{\textbf{Attention behavior of the feature enhancement module.} 
Red lines indicate the top-100 referenced features, indicating that during the self-attention, the query feature (\textcolor{blue}{$\star$}) primarily refers to features from geometrically consistent structures (\eg walls).
}
\vspace{-6mm}
\figlabel{attn_vis}
\end{figure}


\begin{table*}[t]
\centering
\setlength{\tabcolsep}{8pt}
\renewcommand{\arraystretch}{1.15}
\vspace{2mm}
\caption{\textbf{Quantitative localization results on the QEOxford dataset.}
Mean position error (m) and mean orientation error ($^\circ$) are reported. \textbf{Best} and \underline{second} results are in bold and underlined.}

\label{tab:qeoxford}

\begin{tabular}{cl|cccc|c}
\toprule

& \textbf{Method} &
\texttt{15-13-06-37} & \texttt{17-13-26-39} & \texttt{17-14-03-00} & \texttt{18-14-14-42} & Avg. ($\downarrow$) \\
\midrule

& SGLoc~\cite{SGLoc}          & 1.79/1.67 & 1.81/1.76 & 1.33/1.59 & 1.19/1.39 & 1.53/1.60 \\
& LiSA~\cite{LiSA}            & 0.94/1.10 & 1.17/1.21 & 0.84/1.15 & 0.85/1.11 & 0.95/1.14 \\
& LightLoc~\cite{LightLoc}    & 0.82/1.12 & 0.85/1.07 & 0.81/1.11 & \underline{0.82}/1.16 & 0.83/1.12 \\
& GTRLoc~\cite{GTRLoc}       & \underline{0.77/1.02} & \underline{0.77/1.01} & \underline{0.67/1.01} & \textbf{0.80}/\underline{1.07} & \underline{0.75/1.03} \\
& \textbf{ReLoc~(ours)} & \textbf{0.73/0.87} & \textbf{0.64/0.83} & \textbf{0.57/0.85} & \textbf{0.80/0.94} & \textbf{0.69/0.87} \\
\bottomrule
\end{tabular}
\vspace{-3mm}
\end{table*}

\begin{figure*}[t]
\centering
\includegraphics[width=0.95\linewidth]{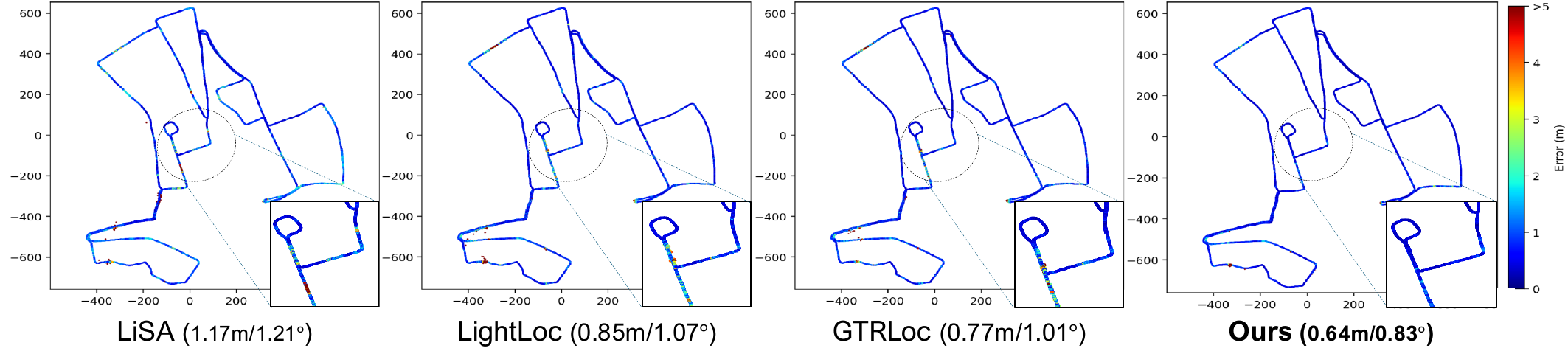}
\vspace{-1mm}
\caption{
\textbf{Qualitative localization result} on \texttt{17-13-26-39} of the QEOxford dataset. Position errors are visualized using a heatmap capped at 5m, where our method exhibits consistently lower errors (dots colored in \textcolor{blue}{blue}) along the trajectory compared to prior approaches.}
\figlabel{qeoxford}
\vspace{-6mm}
\end{figure*}

\subsection{Local feature enhancement module}
\label{sec:local_embedding}

To remain robust against dynamic scenes, we employ a simple yet effective local feature enhancement module ($f_E$) as shown in Fig.~\figref{framework}.
The feature enhancement module consists of a multi-head self-attention layer followed by a feedforward network with normalization and residual connections.
Given the voxel features ($F_P$) from the scene-agnostic backbone~\cite{LightLoc}, the self-attention layer models contextual relationships across all the voxel features within the scan to capture geometrically consistent structures as shown in Fig.~\figref{attn_vis}.
Notably, instead of directly replacing the original features, the attended features are added back to $F_P$ through a residual connection,
\begin{equation}
F_{P} \leftarrow F_{P} + \beta f_E(F_P),
\eqlabel{eq:feature_enhancement}
\end{equation}
where $\beta$ is a hyperparameter.
This residual formulation preserves the underlying geometric cues from the scene-agnostic backbone while injecting context-aware refinements.
In particular, the residual connection not only reinforces geometrically consistent structures across the scan but also attenuates responses from dynamic objects or unstable regions by injecting reliable static context from the self-attention into the noisy local features.
In other words, our proposed module generates offsets that implicitly learn to highlight static features while reducing variance of noisy features by regularizing them against surrounding stable geometry, providing representations beneficial for coordinate regression.

\pparagraph{Loss function.} Following~\cite{LightLoc}, we train the feature enhancement module and MLP regressor using the $L_1$-norm regression loss term ($L_{\text{scr}}$) defined as,
\begin{equation}
L_{\text{scr}} = \sum_{i=1}^{M} \left\| \v p_i^{pred} - \v p_i^{GT} \right\|_1,
\eqlabel{eq:loss_scr}
\end{equation}
where $M$ denotes the number of predicted points.

\subsection{LiDAR localization procedure}
\label{sec:pose_estimation}

Given a voxelized LiDAR scan $S:=\{\v p_i^s \}$, we feed it into the scene-agnostic backbone network, followed by a global embedding module and local feature enhancement module, and concatenated each representation to obtain hierarchical features. 
Finally, we pass them into the MLP regressor to obtain predicted world coordinates $W := \{\v p_i^{pred} \}$.
For 6-DOF pose estimation, we employ the point cloud registration solver~\cite{sc2pcr} with RANSAC, as formulated in Eq.~\eqref{eq:ransac}. Following LightLoc~\cite{LightLoc}, the RANSAC inlier threshold is set to 2m.

\section{Experiments}
\label{sec:experiments}

\subsection{Experimental details}

\pparagraph{Datasets.} Following prior studies~\cite{SGLoc, LiSA, LightLoc, GTRLoc}, we evaluate our method on two public datasets: \textit{Quality Enhanced Oxford (QEOxford)}~\cite{Oxford, SGLoc} and \textit{NCLT}~\cite{NCLT}.
QEOxford~\cite{SGLoc} dataset refines the noisy GPS/INS ground-truth labels of the original Oxford Radar RobotCar dataset~\cite{Oxford}.
As in prior studies, we utilize the left LiDAR stream from a dual Velodyne HDL-32E setup and adopt the same train/test sequences.
NCLT comprises 27 mapping sessions collected by a Segway robot navigating a campus under varying seasonal and illumination conditions, where its ground-truth trajectories are estimated via SLAM.
We consistently follow the identical train/test sequences used in prior studies.
\begin{table*}[t]
\centering
\setlength{\tabcolsep}{8pt}
\renewcommand{\arraystretch}{1.15}
\vspace{2mm}
\caption{\textbf{Quantitative localization results on the NCLT dataset}. Mean position error (m) and mean orientation error ($^\circ$) are reported. \textbf{Best} and \underline{second} results are in bold and underlined.}
\label{tab:nclt}
\begin{tabular}{cc|cccc|c}
\toprule
 & \textbf{Method} &
\texttt{2012-02-12} & \texttt{2012-02-19} & \texttt{2012-03-31} & \texttt{2012-05-26} & Avg.  ($\downarrow$) \\
\midrule

 & SGLoc~\cite{SGLoc}            & 1.20/3.08 & 1.20/3.05 & 1.12/3.28 & 3.48/4.43 & 1.75/3.46 \\
 & LiSA~\cite{LiSA}              & 0.97/\textbf{2.23} & 0.91/\underline{2.09} & 0.87/\underline{2.21} & 3.11/\underline{2.72} & 1.47/\underline{2.31} \\
 & LightLoc~\cite{LightLoc}      & 0.98/2.76 & 0.89/2.51 & 0.86/2.67 & 3.10/3.26 & 1.46/2.80 \\
 & GTRLoc~\cite{GTRLoc}         & \underline{0.95}/2.53 & \underline{0.82}/2.45 & \underline{0.82}/2.52 & \underline{3.01}/2.99 & \underline{1.40}/2.62 \\
& \textbf{ReLoc~(ours)}       & \textbf{0.88}/\underline{2.25} & \textbf{0.76/2.01} & \textbf{0.65/2.18} & \textbf{2.46/2.66} & \textbf{1.19/2.28} \\
\bottomrule
\end{tabular}
\vspace{-4mm}
\end{table*}
\begin{figure*}[t]
\centering
    \centering{\includegraphics[width=0.95\linewidth]{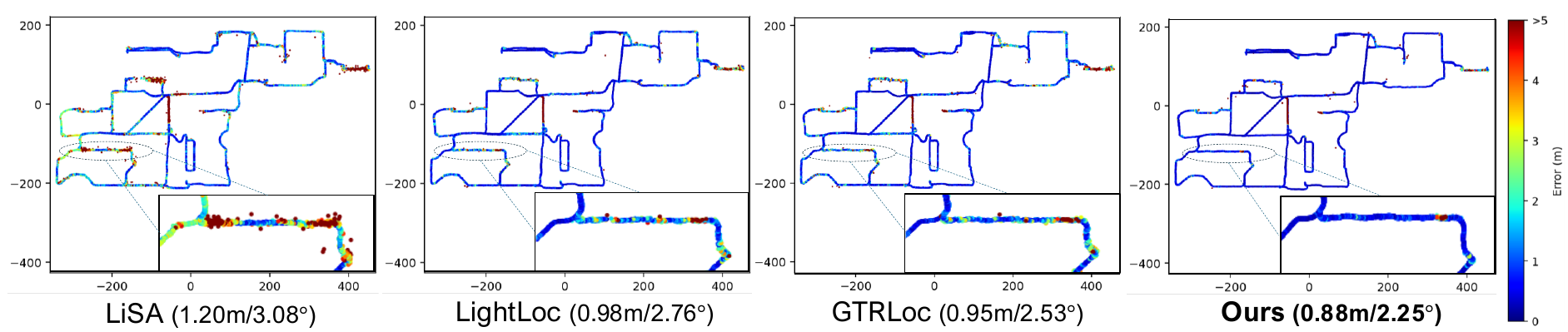}}
    \vspace{-2mm}
    \caption{
    \textbf{Qualitative localization results} on \texttt{2012-02-12} of the NCLT dataset. Position errors are visualized as done in Fig.~\figref{qeoxford}. Our method shows the reduced position error (dots colored in \textcolor{blue}{blue}) along the trajectory compared to prior approaches.}
    \figlabel{nclt}
    \vspace{-4mm}
\end{figure*}

\renewcommand{\arraystretch}{1.2}
\begin{table}[t]
    \centering
    \caption{Ablation study on the QEOxford and NCLT datasets.}
    \vspace{-2mm}
    \begin{tabular}{cc|cc}
        \toprule
         \multirow{2}{*}{\makecell{\textbf{Global}\\\textbf{Embedding}}}  
         & \multirow{2}{*}{\makecell{\textbf{Local Feature}\\\textbf{Enhancement}}} 
         & {QEOxford ($\downarrow$)} & {NCLT ($\downarrow$)} \\
          &   & mean error(m/°) & mean error(m/°)\\
         \hline
          &  & 1.32/1.69 & 2.72/4.18 \\
         \checkmark &  & 0.74/0.94 & 1.25/2.50 \\ 
         \checkmark & \checkmark & \textbf{0.69/0.87} & \textbf{1.19/2.28} \\ \bottomrule
    \end{tabular}
    \vspace{-5mm}
    \label{tab:Ablation_result}
\end{table}

\textbf{Evaluation details.}
For the baseline models, we compare with the LiDAR-based SCR approaches, including the state-of-the-art, GTRLoc~\cite{GTRLoc}. 
For evaluation metrics, we follow~\cite{SGLoc, LiSA, LightLoc, GTRLoc}, reporting mean position and orientation error between predicted poses and ground-truth poses.

\textbf{Implementation details.}
Our framework is implemented using PyTorch and MinkowskiEngine.
We train the global embedding module for 50 epochs with a batch size of 512, and the feature enhancement module~($f_E$) and MLP regressor~($f_H$) for 25 and 30 epochs on QEOxford and NCLT, respectively, with a batch size of 256. 
We use the AdamW optimizer with a one-cycle learning rate schedule, where the learning rate is varied between $5\times10^{-4}$ and $5\times10^{-3}$ during training.
The voxelization is performed with a size of 0.25 m (QEOxford) and 0.3 m (NCLT).
We also set $\beta$ in Eq.~\eqref{eq:feature_enhancement} to 0.1.
During training, we use a workstation with an Intel CPU Xeon Gold 6342 and NVIDIA RTX A6000 GPU.

\subsection{Localization results}

\textbf{Results on the QEOxford dataset.}
As shown in Tab.~\ref{tab:qeoxford}, we evaluate the performance of our proposed method on the QEOxford dataset.
Notably, our method achieves state-of-the-art performance in both mean positional and rotational errors across all four test sequences, outperforming GTRLoc by 8.0\% in positional and 15.5\% in rotational errors on average.
Fig.~\figref{qeoxford} qualitatively illustrates that our method yields trajectories that more closely align with the ground truth, indicating reduced positional errors along the trajectories.

\textbf{Results on the NCLT dataset.}
Evaluation on the NCLT dataset is also reported in Tab.~\ref{tab:nclt}. Similar to the QEOxford, our method achieves state-of-the-art performance on the average of four test sequences among the SCR methods. 
Especially, our method reduces the mean positional and rotational errors by 15\% and 13\% on average compared to GTRLoc.
Fig.~\figref{nclt} also indicates the qualitatively improved localization results of ReLoc compared to previous methods. 

\subsection{Inference time}
The input LiDAR frame rates of the QEOxford and NCLT datasets are 20Hz and 10Hz, respectively.
We evaluate our method on a workstation with an AMD Ryzen 9 7900X 12-Core Processor and a single NVIDIA RTX 4090 GPU.
Our approach achieves average inference speeds of 97Hz on QEOxford and 93Hz on NCLT, demonstrating its capability to satisfy real-time LiDAR localization requirements.

\subsection{Ablation studies}
\pparagraph{Ablation on global embedding module.}
To validate the proposed scene context-aware global embedding module, we conduct ablation studies in Tab.~\ref{tab:Ablation_result}. 
As observed, our global embedding module consistently improves localization performance over the baseline in both datasets.
Furthermore, Fig.~\figref{global_analysis} demonstrates that our proposed global embeddings provide better fine-grained discrimination than classification-based embeddings~\cite{LightLoc,GTRLoc}. 
Specifically, our embeddings eliminate the ambiguities inherent in classification logits within the same cluster by yielding high similarity for nearby scans and lower similarity for distant ones.
Benefiting from this discriminative power, our approach achieves higher localization accuracy than LightLoc~\cite{LightLoc} even when only the global embedding is replaced (Tab.~\ref{tab:Ablation_result}, middle row). 
These results demonstrate the effectiveness of our scene context-aware global representation over the classification logits.

\pparagraph{Ablation on local feature enhancement module.} We also conduct an ablation study on the proposed local feature enhancement module. As shown in Tab.~\ref{tab:Ablation_result}, our proposed module boosts the localization accuracy compared to using naive local features from the scene-agnostic backbone~\cite{LightLoc}, demonstrating that local feature refinement leads to more robust local representations.
Tab.~\ref{tab:ablation_enhancement} also demonstrates that our module improves the median regression error and the inlier ratio of features from not only static regions but also dynamic objects (e.g., cars, trucks). This supports our intuition that injecting consistent context via residual connection reinforces static features, while regularizing noisy dynamic features by referring to the features from the surrounding context-consistent geometry.

\begin{table}[t]
\centering
\footnotesize
\setlength{\tabcolsep}{5pt} 
\renewcommand{\arraystretch}{1.2} 
\caption{Impact of the local feature enhancement module ($f_E$) on the \texttt{17-13-26-39} sequence of the QEOxford dataset.}
\vspace{-2mm}
\label{tab:ablation_enhancement}
\begin{tabular}{lcc|lcc}
\hline
\multicolumn{6}{c}{Median regression error ($L_{scr}$) / Inlier ratio (m / \%) ($\downarrow$ /  $\uparrow$)} \\
\hline
\multicolumn{1}{l|}{Class} & w/o $f_E$ & w/ $f_E$ & \multicolumn{1}{l|}{Class} & w/o $f_E$ & w/ $f_E$ \\
\hline
\multicolumn{1}{l|}{manmade}   & 2.73/55.8 & \textbf{2.24/66.9} & \multicolumn{1}{l|}{terrain} & 2.40/59.6 & \textbf{2.03/68.8} \\ 
\multicolumn{1}{l|}{vegetation} & 2.09/69.8 & \textbf{1.81/78.4} & \multicolumn{1}{l|}{sidewalk} & 2.44/61.5 & \textbf{2.04/71.9} \\
\multicolumn{1}{l|}{car} & 3.53/41.4 & \textbf{3.07/48.1} & \multicolumn{1}{l|}{truck} & 4.71/29.2 & \textbf{4.53/31.9} \\
\hline
\end{tabular}
\vspace{-5mm}
\end{table}

\begin{table}[h]
\centering
\footnotesize
\setlength{\tabcolsep}{2pt} 
\renewcommand{\arraystretch}{1.3} 
\caption{Ablation studies of hyperparameters ($K$ and $\beta$) on the QEOxford dataset. We report mean rotation and translation error of all the test sequences.}
\vspace{-2mm}
\label{tab:ablation_K_beta}
\begin{tabular}{lccc||lccc}
\hline
\multicolumn{1}{l|}{$K$} & \textbf{128}\tiny{(default)} & 256 & 512 & \multicolumn{1}{l|}{$\beta$} & \textbf{0.1}\tiny{(default)} & 0.5 & 1.0 \\
\hline
\multicolumn{1}{l|}{m/$^\circ$} & \textbf{0.69/0.87} & 0.70/0.89 & 0.75/0.88 & \multicolumn{1}{l|}{m/$^\circ$} & \textbf{0.69/0.87} & 0.82/0.87 & 0.92/0.89 \\
\hline
\end{tabular}
\vspace{-2mm}
\end{table}
\pparagraph{Ablation on hyperparameters.} We also investigated the effect of the number of context tokens($K$) and the residual parameter($\beta$ in Eq.~\eqref{eq:feature_enhancement}) in Tab.~\ref{tab:ablation_K_beta}. 
For the context tokens($K$), we empirically observe that increasing $K$ does not lead to a significant improvement in localization accuracy, where $K=128$ achieves the best performance and we set it as a default.
Regarding the residual parameter $\beta$, increasing $\beta$ tends to degrade the translation accuracy. This suggests that excessive feature enhancement can disrupt the original local geometric representation, thereby weakening the balance between context injection and local feature preservation.

\section{CONCLUSION}
Despite recent advancement of SCR-based approaches in LiDAR-based localization, we observed two main limitations in the current architecture, namely (i) use of scene classification-based logits that are incapable of fine-grained discrimination of scene positions, and (ii)  vulnerability to noise induced by dynamic or unstable objects in the scene.
In this work, we presented two simple yet effective modules in SCR each of which addresses these issues.
First, the scene context-aware global embedding module employs a set of learnable context tokens and MLP mixer-based aggregator to capture richer geometric context per scan. Second, the local feature enhancement module adopts self-attention to emphasize stable static structures while mitigating the adverse impact of noisy dynamic features.
Through extensive comparisons and ablation study on two outdoor datasets, we demonstrated both modules synergistically boost the localization performance of LiDAR-based SCR without compromising the real-time inference.
    

\bibliographystyle{ieeetr}
\bibliography{main}
\end{document}